\documentclass[runningheads]{llncs}
\usepackage[T1]{fontenc}
\usepackage{graphicx}
\usepackage{cite} 
\usepackage{hyperref}
\usepackage{microtype}
\usepackage{amsmath,amssymb,amsfonts}
\usepackage{algorithmic}
\usepackage{graphicx}
\usepackage{textcomp}
\usepackage{amsmath}
\usepackage{amssymb}
\usepackage{dsfont}
\usepackage{booktabs}
\usepackage{multirow} 
\usepackage{subcaption}  % For subfigures
\usepackage{caption}     % For captions
\usepackage{float}   
\usepackage[percent]{overpic}
\usepackage{makecell}
\usepackage{arydshln}
\begin{document}
\title{A Low-Resolution Image is Worth 1x1 Words: Enabling Fine Image
Super-Resolution with Transformers and TaylorShift}
\titlerunning{A Low-Resolution Image is Worth 1x1 Words}
\author{Sanath B Nagaraju\inst{1} \and
Brian B Moser\inst{1,2,3} \and
Tobias C Nauen\inst{1,2} \and
Stanislav Frolov\inst{2} \and
Federico Raue\inst{2} \and
Andreas Dengel\inst{1, 2}}
\authorrunning{Nagaraju et al.}
\institute{University of Kaiserslautern-Landau, Germany \and
German Research Center for Artificial Intelligence, Germany \and
Corresponding Author\\
\email{first.last@dfki}}
\maketitle              % typeset the header of the contribution
\begin{abstract}
Transformer-based architectures have recently advanced the image reconstruction quality of super-resolution (SR) models. 
Yet, their scalability remains limited by quadratic attention costs and coarse patch embeddings that weaken pixel-level fidelity. 
We propose TaylorIR, a plug-and-play framework that enforces 1×1 patch embeddings for true pixel-wise reasoning and replaces conventional self-attention with TaylorShift, a Taylor-series-based attention mechanism enabling full token interactions with near-linear complexity. 
Across multiple SR benchmarks, TaylorIR delivers state-of-the-art performance while reducing memory consumption by up to 60\%, effectively bridging the gap between fine-grained detail restoration and efficient transformer scaling.

\end{abstract}

\section{Introduction}
Image super-resolution (SR) aims to recover high-resolution (HR) images from low-resolution (LR) inputs, restoring visual detail that is often critical in domains such as security, medical imaging, and remote sensing \cite{moser2023hitchhiker, moser2024diffusion, xiao2025deep, vu2025comprehensive, li2024systematic}. Despite recent progress, SR remains challenging, particularly in reconstructing high-frequency details that define texture and sharpness \cite{moser2023dwa, liu2019multi, guo2017deep}.

Early SR methods based on convolutional neural networks (CNNs) achieved strong results by learning hierarchical image features \cite{zhang2018image, zhang2018residual, lim2017enhanced}. More recently, transformer-based models have surpassed CNNs by capturing long-range dependencies and richer contextual relationships \cite{vaswani2017attention, dosovitskiy2020image, liang2021swinir, zamir2022restormer, chen2023activating, zhang2022swinfir}. These architectures leverage self-attention mechanisms that model interactions across spatial tokens, leading to superior restoration quality.

However, transformer-based SR still faces two main limitations: high computational cost and loss of fine spatial detail.
First, the \textit{quadratic cost of self-attention} restricts how large an image the model can process at once. To stay feasible, most methods limit attention to small \textit{windows} (e.g., $8{\times}8$ tokens), so each token only interacts with a small neighborhood instead of the entire image.
Second, SR transformers often group pixels into larger \textit{patches} (e.g., $4{\times}4$ or $8{\times}8$), which reduces sequence length but also averages out local variations.
As a result, small windows and large patches restrict long-range reasoning and blur fine details.

\begin{figure}[!t]
    \begin{center}
        \includegraphics[width=.50\linewidth]{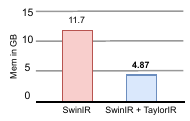}
        \caption{\label{fig:idea}
        Overview of TaylorIR’s impact on image SR. Using 1×1 patch embeddings, TaylorIR models the image at pixel-level resolution. TaylorShift \cite{nauen2024taylorshift} replaces standard self-attention with a Taylor-series-based alternative that maintains full token interaction while reducing memory load.
        }
    \end{center} 
\end{figure}

To overcome these challenges, we introduce TaylorIR, a transformer-based SR framework that enables pixel-level reasoning while remaining efficient. TaylorIR replaces conventional self-attention with TaylorShift \cite{nauen2024taylorshift}, a memory-efficient mechanism derived from Taylor series expansion. It approximates full token-to-token attention with near-linear complexity, making large-scale, per-pixel attention feasible.
For Swin-based SR architectures, we develop TaylorSwinIR, which extends SwinIR to use 1×1 patches and larger attention windows, scaling from 8×8 (64 tokens) to 48×48 (2304 tokens), without prohibitive memory growth. TaylorShift enables this expansion, allowing global context modeling and sharper detail recovery. Compared to baseline SwinIR, TaylorIR achieves higher PSNR and SSIM across multiple benchmarks while cutting memory usage by up to 60\%, as demonstrated empirically in extensive experiments across standard benchmarks and exemplified in \autoref{fig:idea}.

Our main contributions are:
\begin{itemize}
\item \textbf{Pixel-Level Patch Embedding:} We introduce a 1×1 patch embedding strategy that enables transformers to operate directly on pixels, enhancing fine-grained reconstruction.
\item \textbf{TaylorShift Attention:} A Taylor-series-based approximation that achieves global token interaction with near-linear complexity and reduced memory footprint.
\item \textbf{Improved SR Performance and Efficiency:} TaylorSwinIR consistently outperforms current SR transformers across standard datasets, offering a stronger balance between reconstruction quality and computational efficiency.
\end{itemize}
\section{Background: Self-Attention}
% Transformer-based Image Super-Resolution

%This section reviews the standard self-attention mechanism~\cite{vaswani2017attention}.

%\subsection{Classical Self-Attention}

In transformers~\cite{vaswani2017attention, DBLP:journals/corr/abs-2010-11929}, the attention mechanism determines how strongly each token in a sequence relates to every other token.
For a sequence of length N, standard self-attention computes pairwise interactions between all tokens, enabling global context modeling but at a quadratic computational cost.

The process involves three main steps:

\begin{enumerate}
\item \textbf{Query–Key Similarity:} Compute pairwise affinities between queries Q and keys K:
\begin{equation}
A = QK^\top \in \mathbb{R}^{N \times N}.
\end{equation}
\item \textbf{Softmax Normalization:} Convert similarities into normalized weights:
\begin{equation} \label{eq:softmax_activation}
    \text{softmax}(A)_{ij} = \frac{\exp(A_{ij})}{\sum_k \exp(A_{ik})}.
\end{equation}

\item \textbf{Value Aggregation:} Weight the value matrix $V$ using the attention scores:
\begin{equation}
    Y = \text{softmax}(A)V.
\end{equation}
\end{enumerate}

%remove this
%The total Floating Point Operations (FLOPS) for the attention mechanism is
%\begin{equation}
%\text{FLOPS} = 2N^2d + 4N^2.
%\end{equation}
Overall, the time complexity of classical self-attention is $\mathcal{O}(N^2 d)$, which becomes computationally expensive, particularly for long sequences.

\section{TaylorIR}
TaylorIR is a simple drop-in recipe to make transformer SR both \emph{finer} and \emph{leaner}. 
It has two parts:
(i) TaylorShift attention to keep long-range interactions affordable in time and memory, and
(ii) pixel-wise patch embedding ($1{\times}1$) to let the model reason at the level of individual pixels.

\subsection{TaylorShift Attention}
TaylorShift replaces the softmax in attention with a low-order Taylor expansion. It keeps the expressiveness of full token-to-token interactions while reducing runtime and memory.

\subsubsection{Direct TaylorShift}
Let $A = QK^\top \in \mathbb{R}^{N \times N}$. Direct TaylorShift uses a second-order Taylor approximation of the exponential and normalizes it to a valid distribution:
\begin{equation}
\label{eq:taylor-sm}
\mathrm{T\text{-}SM}(A)_{ij} \;=\;
\frac{1 + A_{ij} + \tfrac{1}{2}A_{ij}^2}{\sum\nolimits_{k}\big(1 + A_{ik} + \tfrac{1}{2}A_{ik}^2\big)}.
\end{equation}
The output is then $Y=\mathrm{T\text{-}SM}(A)V$. This avoids exponentials and is slightly faster than softmax, but it still has the same asymptotic cost, $\mathcal{O}(N^2 d)$, because it forms and uses the $N{\times}N$ matrix.

\subsubsection{Efficient TaylorShift}
For long sequences, we reorder the computation so that normalization happens \emph{after} mixing with $V$. Define
\begin{align}
\label{eq:eff-nom-den}
Y_{\text{nom}} &= \big[\,\mathbf{1} + QK^\top + \tfrac{1}{2}(QK^\top)^{\odot 2}\,\big]\,V,\\
Y_{\text{den}} &= \big[\,\mathbf{1} + QK^\top + \tfrac{1}{2}(QK^\top)^{\odot 2}\,\big]\,\mathds{1}_N,\nonumber
\end{align}
and return the elementwise normalized result $Y = Y_{\text{nom}} \oslash Y_{\text{den}}$, where $\mathds{1}_N$ is the length-$N$ vector of ones and $\odot$ / $\oslash$ denote Hadamard power and division.

To avoid building $(QK^\top)^{\odot 2}$ explicitly, TaylorShift expands the squared term with a tensor-like operator $\boxtimes : \mathbb{R}^{N\times d}\times \mathbb{R}^{N\times d}\to \mathbb{R}^{N\times d^2}$,
\begin{equation}
\label{eq:box}
(QK^\top)^{\odot 2}V \;=\; (Q\boxtimes Q)\,\big(K\boxtimes K\big)^\top V,
\end{equation}
so the constant and linear parts run in $\mathcal{O}(Nd^2)$ and the squared part in $\mathcal{O}(Nd^3)$. In practice, Efficient TaylorShift is preferable once $N$ is large (i.e., when $Nd^3 \ll N^2 d$ no longer holds).

\begin{figure*}[!t]
    \centering
    \includegraphics[width=\textwidth]{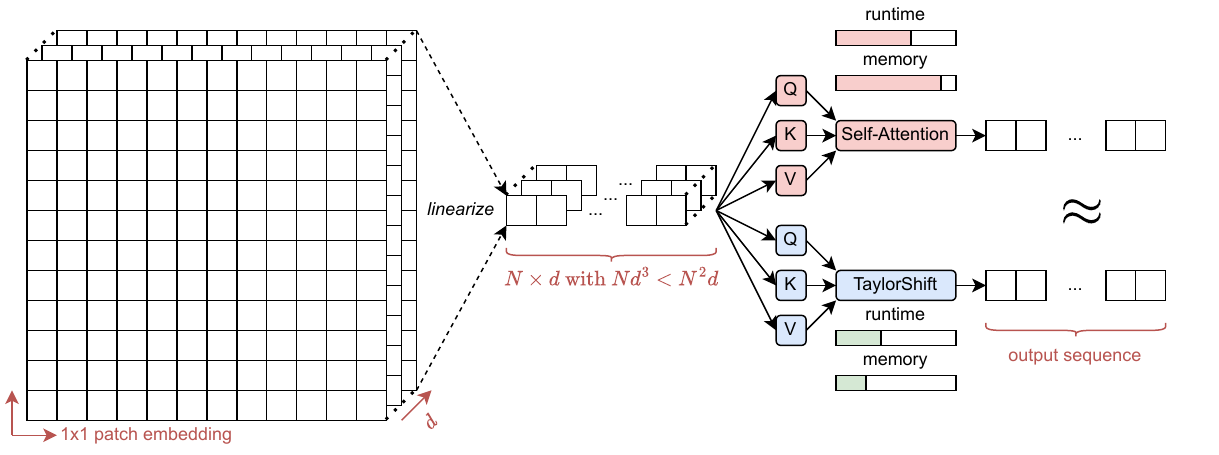}
    \caption{\label{fig:taylorswinir}
    TaylorIR has two pieces: (\emph{left}) pixel-wise ($1{\times}1$) patch embedding and (\emph{right}) TaylorShift attention in place of windowed softmax attention. Together, they enable long-range context with lower memory and stable runtime.}
\end{figure*}

\subsection{Pixel-Wise Patch Embedding}
Most ViT-based SR architectures reduce sequence length by partitioning the input into $p{\times}p$ patches, typically with $p\in\{4,8\}$. While this lowers computational cost, it implicitly enforces spatial smoothness within each patch and limits fine-grained control. TaylorIR instead adopts a degenerate patch size of $p{=}1$, embedding each pixel as an independent token. An image $\mathbf{x}\in\mathbb{R}^{H\times W}$ is thus reshaped into a sequence $\{\mathbf{x}_i\}_{i=1}^{N}$ with $N{=}HW$. This pixel-wise embedding exposes the full-resolution spatial manifold to the network, removes the hidden downsampling effect of large patches, and aligns naturally with the TaylorShift operator, which maintains tractable attention over long spatial windows.

\subsection{Integrating TaylorIR into SwinIR}
\label{sec:TaylorSwinIR}
To evaluate compatibility with hierarchical backbones, we integrate TaylorIR into SwinIR, yielding TaylorSwinIR. While SwinIR already employs pixel embeddings, its attention is confined to non-overlapping $w{\times}w$ windows, with $w{=}8$ (i.e., $64$ tokens). We extend the window size to $w{=}48$ (i.e., $2304$ tokens) and replace the original window attention $\mathcal{A}_{\text{win}}$ with the TaylorShift attention $\mathcal{A}_{\text{TS}}$. Formally,
\begin{equation}
\mathbf{y} = \mathcal{A}_{\text{TS}}(\mathbf{Q}, \mathbf{K}, \mathbf{V}),
\end{equation}
where $\mathcal{A}_{\text{TS}}$ can operate either in its direct form for short sequences or in its efficient variant for long ones, depending on $w^2$. This substitution broadens the effective receptive field while keeping the asymptotic cost near-linear in $N$. Empirically, TaylorSwinIR preserves or improves reconstruction quality, reduces VRAM consumption at large window sizes, and remains fully drop-in compatible with SwinIR-style architectures.

\begin{table*}[t!]\scriptsize
\center
\begin{center}
\caption{Quantitative comparison (average PSNR/SSIM) of our proposed TaylorSwinIR model, which applies TaylorIR to SwinIR, against state-of-the-art methods for classical image SR on benchmark datasets (Set5, Set14, BSD100, Urban100, and Manga109). As a result, TaylorSwinIR improves the performance of its underlying architecture, SwinIR, on most test sets.}%
\label{tab:sr_results}
\resizebox{\textwidth}{!}{%
\begin{tabular}{lcccccccccccc}
\toprule
\multirow{2}{*}{Method} & \multirow{2}{*}{Scale} & \multirow{2}{*}{\makecell{Training\\Dataset}} &  \multicolumn{2}{c}{Set5~\cite{Set5}} &  \multicolumn{2}{c}{Set14~\cite{Set14}} &  \multicolumn{2}{c}{BSD100~\cite{BSD100}} &  \multicolumn{2}{c}{Urban100~\cite{Urban100}} &  \multicolumn{2}{c}{Manga109~\cite{Manga109}}  
\\
\cmidrule(r){4-5} \cmidrule(lr){6-7} \cmidrule(lr){8-9} \cmidrule(lr){10-11} \cmidrule(l){12-13}
&  &  & PSNR & SSIM & PSNR & SSIM & PSNR & SSIM & PSNR & SSIM & PSNR & SSIM 
\\
\midrule

RCAN~\cite{zhang2018rcan} & $\times$2 & DIV2K %
& {38.27}
& {0.9614}
& {34.12}
& {0.9216}
& {32.41}
& {0.9027}
& {33.34}
& {0.9384}
& {39.44}
& {0.9786}
\\  
SAN~\cite{dai2019SAN} & $\times$2 & DIV2K %
& {38.31}
& {0.9620}
& {34.07}
& {0.9213}
& {32.42}
& {0.9028}
& {33.10}
& {0.9370}
& {39.32}
& {0.9792}
\\
IGNN~\cite{zhou2020IGNN} & $\times$2 & DIV2K %
& {38.24}
& {0.9613}
& {34.07}
& {0.9217}
& {32.41}
& {0.9025}
& {33.23}
& {0.9383}
& {39.35}
& {0.9786}
\\
HAN~\cite{niu2020HAN} & $\times$2 & DIV2K %
& {38.27}
& {0.9614}
& {34.16}
& {0.9217}
& {32.41}
& {0.9027}
& {33.35}
& {0.9385}
& {39.46}
& {0.9785}  
\\ 
NLSA~\cite{mei2021NLSA} & $\times$2 & DIV2K %
& 38.34 
& 0.9618 
& 34.08 
& {0.9231}
& 32.43 
& 0.9027 
& {33.42}
& {0.9394}
& {39.59}
& 0.9789
\\
SwinIR~\cite{liang2021swinir}  & $\times$2  & DIV2K
& {38.35}
& {0.9620}
& {34.14}
& {0.9227}
& {32.44}
& {0.9030}
& {33.40}
& {0.9393}
& {39.60}
& {0.9792}
\\
\textbf{TaylorSwinIR} (ours) & $\times 2$ & DIV2K & \textbf{38.46} & \textbf{0.9627} & \textbf{34.20} & \textbf{0.9235} & \textbf{32.49} & \textbf{0.9045} & \textbf{33.71} & \textbf{0.9417} & \textbf{39.70} & \textbf{0.9795} \\

\cmidrule(r){1-3} \cmidrule(lr){4-5} \cmidrule(lr){6-7} \cmidrule(lr){8-9} \cmidrule(lr){10-11} \cmidrule(l){12-13}

RCAN~\cite{zhang2018rcan}& $\times$3   & DIV2K
& {34.74}
&{0.9299}
& {30.65}
& {0.8482}
& {29.32}
& {0.8111}
& {29.09}
& {0.8702}
& {34.44}
&{0.9499}
\\
SAN~\cite{dai2019SAN} & $\times$3   & DIV2K
& {34.75}
& {0.9300}
& {30.59}
& {0.8476}
& {29.33}
& {0.8112}
& {28.93}
& {0.8671}
& {34.30}
& {0.9494}
\\
IGNN~\cite{zhou2020IGNN} & $\times$3  & DIV2K
& {34.72}
& {0.9298}
& {30.66}
& {0.8484}
& {29.31}
& {0.8105}
& {29.03}
& {0.8696}
& {34.39}
& {0.9496}
\\

HAN~\cite{niu2020HAN}  & $\times$3   & DIV2K
& {34.75}
& {0.9299}
& {30.67}
& {0.8483}
& {29.32}
& {0.8110}
& {29.10}
& {0.8705}
& {34.48}
& {0.9500}
\\
NLSA~\cite{mei2021NLSA} & $\times$3  & DIV2K
& 34.85 
& 0.9306 
& 30.70 
& 0.8485 
& 29.34 
& 0.8117 
& {29.25}
& {0.8726}
& 34.57 
& 0.9508
\\
SwinIR~\cite{liang2021swinir}  & $\times$3  & DIV2K
& {34.89}
& {0.9312}
& {30.77}
& {0.8503}
& {29.37}
& {0.8124}
& {29.29}
& {0.8744}
& {34.74}
& {0.9518}
\\
\textbf{TaylorSwinIR} (ours) & $\times 3$ & DIV2K & \textbf{34.94} & \textbf{0.9317} & \textbf{30.79} & \textbf{0.8507} & \textbf{29.39} & \textbf{0.8145} & \textbf{29.47} & \textbf{0.8772} & \textbf{34.76} & \textbf{0.9521} \\

\cmidrule(r){1-3} \cmidrule(lr){4-5} \cmidrule(lr){6-7} \cmidrule(lr){8-9} \cmidrule(lr){10-11} \cmidrule(l){12-13}

RCAN~\cite{zhang2018rcan}& $\times$4  & DIV2K
& {32.63}
& {0.9002}
& {28.87}
&{0.7889}
& {27.77}
& {0.7436}
&{26.82}
& {0.8087}
&{31.22}
& {0.9173}
\\ 
SAN~\cite{dai2019SAN} & $\times$4  & DIV2K
& {32.64}
& {0.9003}
& {28.92}
& {0.7888}
& {27.78}
& {0.7436}
& {26.79}
& {0.8068}
& {31.18}
& {0.9169}
\\
IGNN~\cite{zhou2020IGNN}  & $\times$4  & DIV2K
& {32.57}
& {0.8998}
& {28.85}
& {0.7891}
& {27.77}
& {0.7434}
& {26.84}
& {0.8090}
& {31.28}
& {0.9182}
\\

HAN~\cite{niu2020HAN}  & $\times$4  & DIV2K
& {32.64}
& {0.9002}
& {28.90}
& {0.7890}
& {27.80}
& {0.7442}
& {26.85}
& {0.8094}
& {31.42}
& {0.9177}
\\
NLSA~\cite{mei2021NLSA} & $\times$4 & DIV2K
& 32.59 
& 0.9000 
& 28.87 
& 0.7891 
& 27.78 
& 0.7444 
& {26.96}
& {0.8109}
& 31.27 
& 0.9184
\\
SwinIR~\cite{liang2021swinir}  & $\times$4  & DIV2K
& {32.72}
& {0.9021}
& {28.94}
& {0.7914}
& {27.83}
& {0.7459}
& \textbf{27.07}
& \textbf{0.8164}
& \textbf{31.67}
& \textbf{0.9226}
\\
\textbf{TaylorSwinIR} (ours) & $\times 4$ & DIV2K & \textbf{32.74} & \textbf{0.9023} & \textbf{28.99} & \textbf{0.7917} & \textbf{27.83} & \textbf{0.7477} & 27.06 & 0.8156 & 31.57 & 0.9225 \\
\bottomrule
\end{tabular}
}
\end{center}
\vspace{1mm}
\end{table*}

\section{Experiments}
% Short Intro, omit for now
This section introduces our experiments conducted to assess the effectiveness of TaylorIR by applying it to SwinIR. 
First, we describe the setup and analyze the impact of the integration of TaylorIR.
Next, we demonstrate how TaylorIR improves state-of-the-art transformer-based models on classical benchmarks and finally show how TaylorIR improves the contextual scope.

\subsection{Experimental Setup}
We train all models on the DIV2K dataset~\cite{agustsson2017ntire}, following the standard protocol of extracting $192\times192$ RGB sub-images for training. 
Evaluation is performed on the benchmark datasets Set5~\cite{Set5}, Set14~\cite{Set14}, BSDS100~\cite{BSD100}, Manga109~\cite{Manga109}, and Urban100~\cite{Urban100}. 
We report results for upscaling factors of $\times2$, $\times3$, and $\times4$, using PSNR and SSIM on the luminance channel of the YCbCr space for fair comparison with prior work.

\subsection{State-of-the-Art Comparison}

To demonstrate the effectiveness of TaylorSwinIR in classical image SR benchmarks, we compare it against prominent state-of-the-art models including RCAN \cite{zhang2018rcan}, RRDB \cite{wang2018esrgan}, SAN \cite{dai2019second}, IGNN \cite{zhou2020cross}, HAN \cite{niu2020single}, NLSA \cite{mei2021image}, and the baseline SwinIR model \cite{liang2021swinir}. Evaluations were conducted across multiple datasets (Set5, Set14, BSD100, Urban100, and Manga109) and scaling factors ($\times$2, $\times$3, and $\times$4) to ensure comprehensive analysis.
Results are shown in \autoref{tab:sr_results}.

TaylorSwinIR consistently outperforms SwinIR across most datasets, particularly at lower scaling factors ($\times2$ and $\times3$), highlighting the advantage of pixel-wise embeddings combined with TaylorShift attention. 
On Urban100 at $\times2$, TaylorSwinIR achieves a PSNR gain of \textbf{+0.31 dB} over SwinIR, demonstrating its improved ability to reconstruct fine, high-frequency structures common in urban imagery. 
Similarly, on Manga109 at $\times2$, it reaches \textbf{39.70 dB} PSNR versus \textbf{39.60 dB} for SwinIR, reflecting superior detail preservation in dense line-art textures. 
Across all benchmarks and scales, TaylorSwinIR yields PSNR improvements ranging from \textbf{+0.02} to \textbf{+0.31 dB}, accompanied by consistent SSIM gains, confirming the stability of its enhancements.
At $\times4$ magnification, TaylorSwinIR maintains parity with SwinIR on most datasets, indicating that the efficiency-oriented TaylorShift mechanism does not compromise reconstruction fidelity at higher scales. 

In summary, TaylorSwinIR establishes a new benchmark among transformer-based SR models by jointly delivering high perceptual fidelity and computational efficiency. 
Beyond its performance gains, the core advantage of \textbf{TaylorIR} lies in its \emph{plug-and-play} design: it can be seamlessly integrated into existing attention-based architectures without architectural modifications. 
This modularity enables straightforward adoption across diverse SR frameworks. 
Together, these results position TaylorIR as a flexible and scalable foundation for future high-quality image SR systems.

\begin{figure*}[!t]
  \centering
  \begin{overpic}[width=\linewidth]{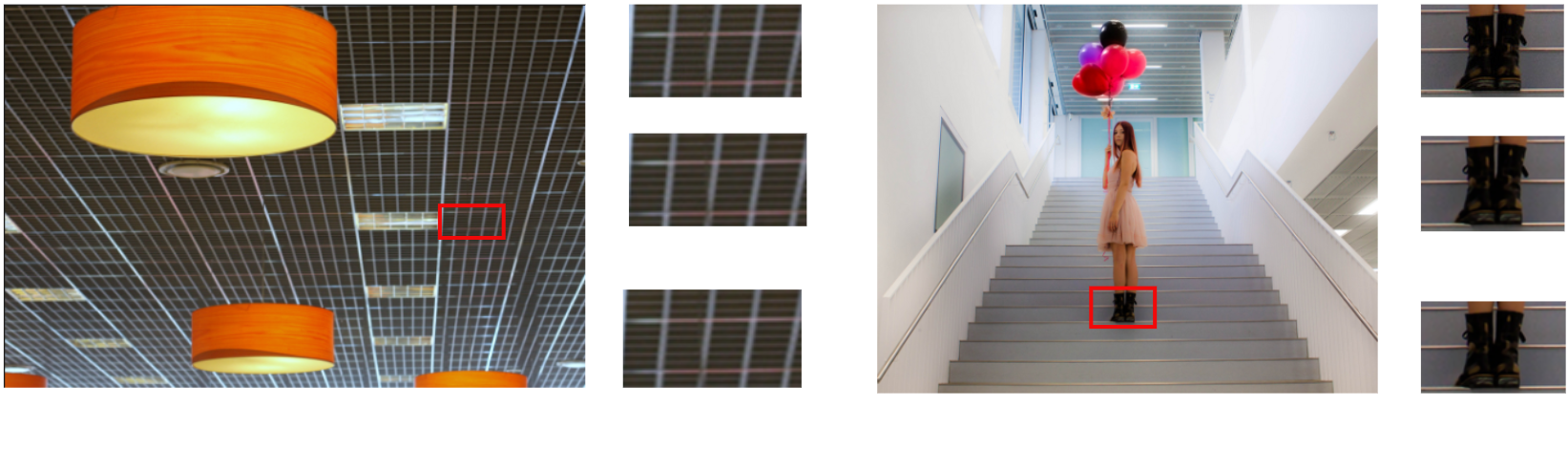}
     \put(11,0){\footnotesize{Urban100 ($\times$2): img\_044}}
     \put(65,0){\footnotesize{Urban100 ($\times$2): img\_009}}
     \put(44,22){\footnotesize{HR}}
     \put(42,13){\footnotesize{SwinIR}}
     \put(43,3){\footnotesize{Ours}}
     \put(94,22){\footnotesize{HR}}
     \put(92,13){\footnotesize{SwinIR}}
     \put(93,3){\footnotesize{Ours}}
  \end{overpic}
  \vspace{1mm}
  \caption{\textbf{Visual comparison on $2{\times}$ SR.} 
  Red boxes in the HR images mark the regions shown for comparison. 
  On \textbf{(left)} Urban100 image \texttt{img\_044}, TaylorSwinIR achieves 44.87\,dB / 0.9921 versus SwinIR’s 44.13\,dB / 0.9905. 
  On \textbf{(right)} Urban100 image \texttt{img\_009}, TaylorSwinIR achieves 42.65\,dB / 0.9834 versus SwinIR’s 42.20\,dB / 0.9834. }
  \label{fig:scale2}
  \vspace{1pt}
\end{figure*}

\subsection{Qualitative Results}
\begin{figure*}[!t]
  \centering
  \begin{overpic}[width=\linewidth]{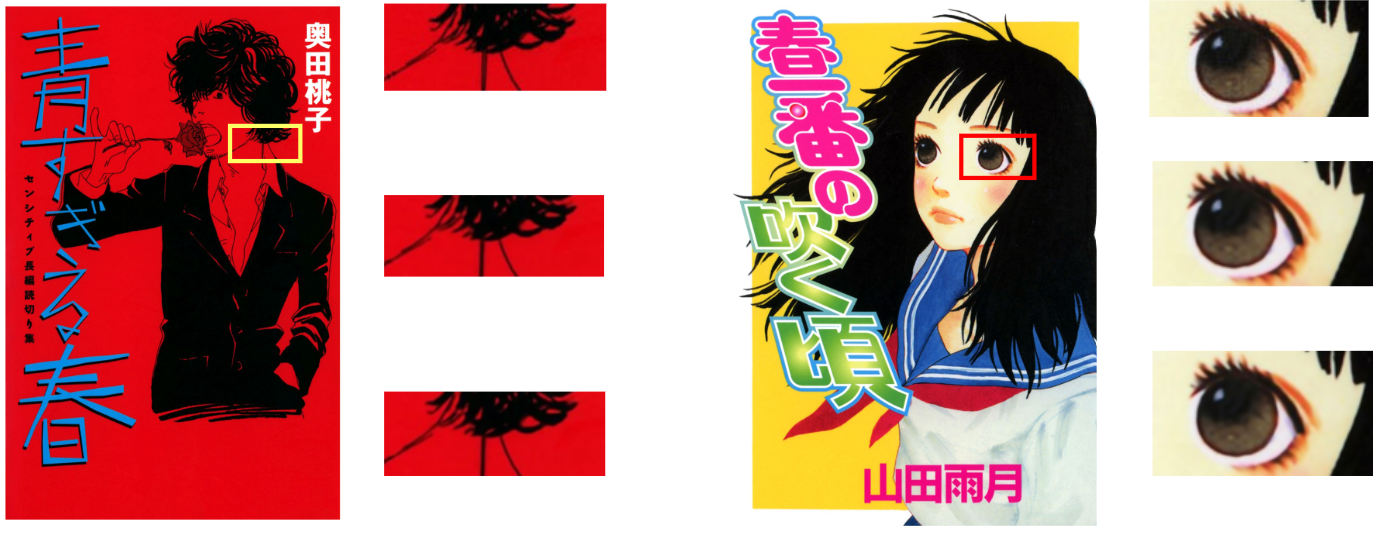}
     \put(3,0){\footnotesize{Manga109 ($\times$3): AosugiruHaru}}
     \put(54,0){\footnotesize{Manga109 ($\times$3): HaruichibanNoFukukoro}}
     \put(34,32){\footnotesize{HR}}
     \put(33,17){\footnotesize{SwinIR}}
     \put(33,4){\footnotesize{Ours}}
     \put(90,30){\footnotesize{HR}}
     \put(88,17){\footnotesize{SwinIR}}
     \put(89,4){\footnotesize{Ours}}
  \end{overpic}
  \vspace{1mm}
  \caption{\textbf{Visual comparison on $3{\times}$ SR.} 
  Yellow and red boxes in the HR images indicate the regions shown for comparison. 
  On \textbf{(left)} Manga109 image \texttt{AosugiruHaru}, TaylorSwinIR achieves \textbf{44.30\,dB / 0.9898} versus SwinIR’s 44.15\,dB / 0.9896. 
  On \textbf{(right)} image \texttt{HaruichibanNoFukukoro}, TaylorSwinIR attains \textbf{42.79\,dB / 0.9872} compared to SwinIR’s 42.47\,dB / 0.9870.}
  \label{fig:scale3}
  \vspace{1pt}
\end{figure*}

\begin{figure*}[!t]
  \centering
  \begin{overpic}[width=\linewidth]{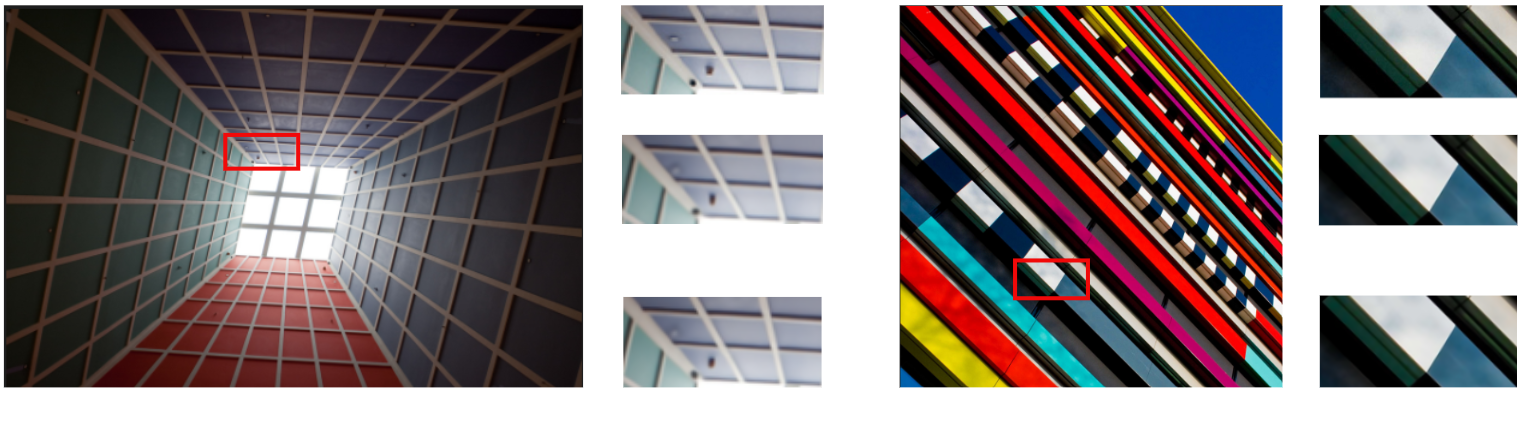}
     \put(10,-2){\footnotesize{Urban100 ($\times$4): img\_090}}
     \put(66,-2){\footnotesize{Urban100 ($\times$4): img\_081}}
     \put(46,21){\footnotesize{HR}}
     \put(44,11){\footnotesize{SwinIR}}
     \put(45,1){\footnotesize{Ours}}
     \put(92,20){\footnotesize{HR}}
     \put(89,11){\footnotesize{SwinIR}}
     \put(91,1){\footnotesize{Ours}}
  \end{overpic}
  \vspace{1mm}
  \caption{\textbf{Visual comparison on $4{\times}$ SR.} 
  Red boxes in the HR images indicate the cropped regions for comparison. 
  On \textbf{(left)} Urban100 image \texttt{img\_090}, TaylorSwinIR achieves \textbf{40.54\,dB / 0.9836} versus SwinIR’s 40.34\,dB / 0.9832. 
  On \textbf{(right)} image \texttt{img\_081}, TaylorSwinIR reaches \textbf{39.81\,dB / 0.9807} compared to SwinIR’s 39.80\,dB / 0.9802. }
  \label{fig:scale4}
  \vspace{1pt}
\end{figure*}

% Insert Visual Comparison (SwinIR vs Ours)
% Insert Attention Map Visualizations for same images (Interpret differences)
In \autoref{fig:scale2}, \autoref{fig:scale3} and \autoref{fig:scale4}, we present visual comparisons across different scaling factors ($\times$2, $\times$3, and $\times$4) between the benchmark model SwinIR and our proposed TaylorSwinIR approach. 
Notably, in \autoref{fig:scale3}, TaylorSwinIR demonstrates clear improvements at the $\times$3 scale for the Manga109 image, \textit{MiraiSan}, where it produces significantly sharper and more detailed results.

TaylorSwinIR consistently achieves subjectively comparable or superior image quality relative to SwinIR across all tested scales. 
These gains are attributed to the expanded receptive field, which enhances the model’s ability to capture richer contextual information. 
Additionally, TaylorSwinIR demonstrates optimized memory efficiency, which not only supports high-quality image reconstruction but also makes it more computationally viable. 
This combination of improved super-resolution quality and reduced memory consumption highlights TaylorSwinIR as an effective and resource-efficient solution for detailed image enhancement.

\subsection{Analysis of Self-Attention vs.\ TaylorShift}
We analyze how large window sizes and attention mechanisms affect runtime and memory. 
To control for context, we fix the window at $48{\times}48$ and compare \emph{SwinIR (fine)} with \emph{TaylorSwinIR} (TaylorShift). 
For reference, we also report the throughput of the default \emph{SwinIR (original)} with $8{\times}8$ windows, noting that it uses a different window size.

\begin{figure}[t]
    \centering
    % Use the PDF in your repo (preferred for CVPR); PNG is fine for drafts.
    % Replace the filename if you relocate the asset.
    \includegraphics[width=\linewidth]{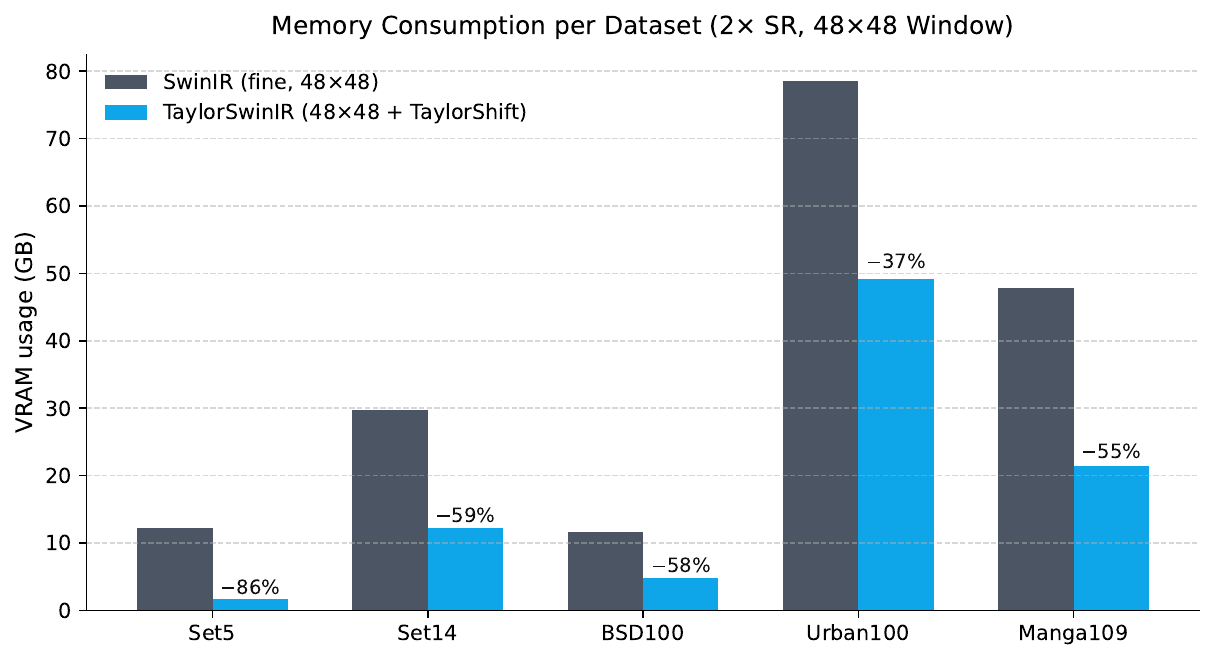}
    \caption{\textbf{Memory consumption} for $2{\times}$ SR at $48{\times}48$ windows.
    TaylorSwinIR (48${\times}$48 + TaylorShift) substantially reduces VRAM usage compared to SwinIR (fine, 48${\times}$48) across datasets, with per-dataset savings annotated above the bars (percentage reduction vs.\ SwinIR~(fine)).}
    \label{fig:vram_analysis}
\end{figure}

\subsubsection{Impact on Memory Efficiency (VRAM)}
At the same $48{\times}48$ window size, \emph{TaylorSwinIR} consistently uses less VRAM than \emph{SwinIR (fine)} across datasets, with observed reductions ranging from roughly $37\%$ to $85\%$, as shown in \autoref{fig:vram_analysis}. 
For example, on Urban100, memory drops from $78.5$\,GB to $49.1$\,GB. 
These savings enable large-window inference under tighter hardware budgets and make wide-context SR more practical at scale.

\subsubsection{Impact on Computational Complexity (Throughput)}
With $8{\times}8$ windows, \emph{SwinIR (original)} reaches $6.12$ images/s due to small local windows and low attention cost. 
At matched large windows ($48{\times}48$), \emph{SwinIR (fine)} runs at $0.123$ images/s, while \emph{TaylorSwinIR} attains $0.124$ images/s under the same setting. 
Although the absolute gain is modest at $48{\times}48$, TaylorShift’s scaling becomes increasingly favorable as sequence length grows, mitigating the quadratic sensitivity of standard windowed self-attention to window area.

\subsubsection{Practical Trade-Offs}
Large windows expand contextual reasoning but stress throughput and memory. 
TaylorShift preserves most of the runtime characteristics at $48{\times}48$ while substantially lowering the memory footprint, offering a more deployable path to large-context SR. 
In practice, this plug-and-play swap of the window attention with TaylorShift amortizes costs as resolution and context grow, without altering the backbone architecture.

\begin{figure*}[!h]
  \centering
  \begin{overpic}[width=\linewidth]{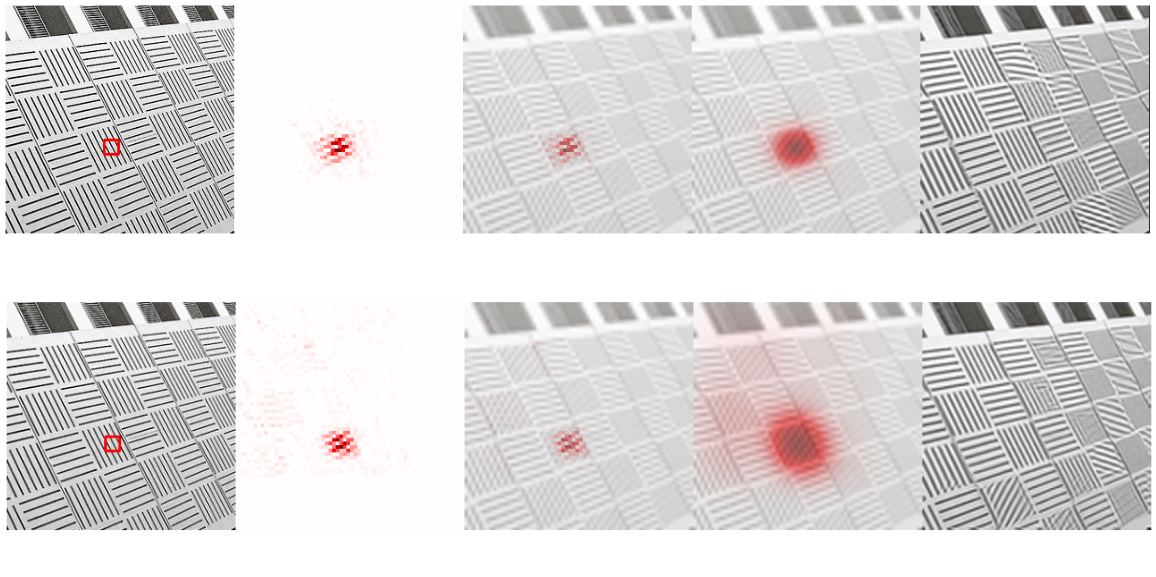}
     \put(9,28){\footnotesize{(a)}}
     \put(29,28){\footnotesize{(b)}}
     \put(49,28){\footnotesize{(c)}}
     \put(69,28){\footnotesize{(d)}}
     \put(89,28){\footnotesize{(e)}}
     \put(0,25){\footnotesize{\text{SwinIR}: Diffusion Index = 5.12; $\text{PSNR/SSIM} = 24.31\text{dB}/0.8626$}}

     \put(9,2){\footnotesize{(a)}}
     \put(29,2){\footnotesize{(b)}}
     \put(49,2){\footnotesize{(c)}}
     \put(69,2){\footnotesize{(d)}}
     \put(89,2){\footnotesize{(e)}}
     \put(0,-1){\footnotesize{\text{TaylorSwinIR}: Diffusion Index = \textbf{14.87}; $\text{PSNR/SSIM} = \textbf{24.95}\text{dB}/\textbf{0.8723}$}}
     
  \end{overpic}

  \vspace{2.5em} % Adjust vertical space before the line as needed
  \rule{\linewidth}{0.5pt} % Horizontal line (width of line, thickness)
  \vspace{2.5em} % Adjust vertical space after the line as needed

  \begin{overpic}[width=\linewidth]{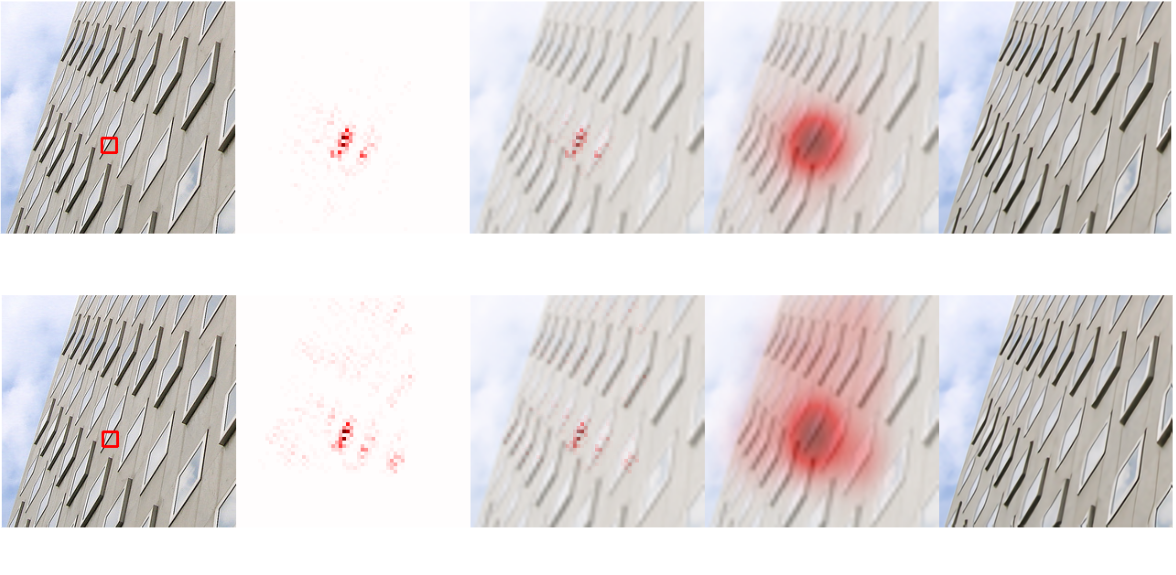}
     \put(9,27){\footnotesize{(a)}}
     \put(29,27){\footnotesize{(b)}}
     \put(49,27){\footnotesize{(c)}}
     \put(69,27){\footnotesize{(d)}}
     \put(89,27){\footnotesize{(e)}}
     \put(0,24){\footnotesize{\text{SwinIR}: Diffusion Index = 12.52; $\text{PSNR/SSIM} = 28.96\text{dB}/0.9143$}}

     \put(9,2){\footnotesize{(a)}}
     \put(29,2){\footnotesize{(b)}}
     \put(49,2){\footnotesize{(c)}}
     \put(69,2){\footnotesize{(d)}}
     \put(89,2){\footnotesize{(e)}}
     \put(0,-1){\footnotesize{\text{TaylorSwinIR}: Diffusion Index = \textbf{13.99}; $\text{PSNR/SSIM} = \textbf{29.42}\text{dB}/\textbf{0.9191}$}}
     
  \end{overpic}
  \vspace{0.5pt}
  \caption{Comparison of Local Attention Maps (LAM) \cite{gu2021interpreting} between SwinIR and TaylorSwinIR for 2$\times$ \textbf{(top)} and 4$\times$ \textbf{(bottom)} image SR. 
  From left to right, the images show: (a) The original image with a selected region highlighted (red boxes), (b) the LAM visualization, (c) the LAM result overlaid on the input image, (d) the informative area overlaid on the input image, and (e) the final super-resolved output image. Overall, TaylorIR leads to a higher diffusion index, which shows its extended contextual scope and improved image quality metrics compared to SwinIR without TaylorIR.}
  \label{fig:lam}
  \vspace{1pt} % Adjust spacing as needed
\end{figure*}

\subsection{Contextual Scope}
\autoref{fig:lam} demonstrates the main advantages of TaylorIR underlying its improved reconstruction quality: the extended contextual scope.
We demonstrate this by visualizing Local Attention Maps (LAM) \cite{gu2021interpreting} at a scaling of 2$\times$ and 4$\times$. 
Unlike the unmodified SwinIR, which is restricted by its smaller window field, TaylorIR enables a broader diffusion of contextual information while keeping memory consumption low, as shown in the previous Sections. 
This results in an expanded Local Attention Map, where a higher diffusion index showcases a greater spread of information used for reconstruction across relevant areas within the image.

\section{Limitations \& Future Work}
While TaylorIR demonstrates enhanced memory efficiency and performance in image super-resolution, several limitations remain. 
Although TaylorShift attention significantly reduces computational complexity, the computation still face substantial memory demands for extremely high-resolution images.

Future work could explore hybrid models that combine the strengths of convolutional and transformer architectures to improve performance across diverse image types. 
Another promising direction would be the optimization of TaylorIR for real-time applications on edge devices by incorporating quantization and model compression techniques.

\section{Conclusion}
In this work, we introduced TaylorIR, a method to extend transformer-based SR models by replacing the standard self-attention mechanism with TaylorShift attention and enabling long sequences, primarily through 1$\times$1 patch embeddings, to be processed faster and significantly more memory efficient.
More specifically, our proposed method allowed us to scale up SwinIR with a 48×48 window, resulting in a sequence length of 2304, capturing more global context, which leads to enhanced image quality while maintaining memory efficiency - a feat unattainable with traditional attention mechanisms.
TaylorSwinIR, the TaylorIR-extended SwinIR architecture, achieves superior PSNR and SSIM scores across classical test benchmarks like Set5, Set14, BSD100, Urban100, and Manga109, surpassing state-of-the-art models, including SwinIR. 
This reduced memory footprint is particularly meaningful for deploying SR transformers in resource-constrained environments, bringing high-quality SR closer to real-time applications.

\section{Societal Impact}

While using long sequences in image SR enables enhanced detail and quality, it also introduces significant computational complexity and memory consumption. 
Although our proposed method, TaylorIR, reduces these demands, the resources required remain substantial. 
Continued improvements in efficiency are crucial to minimize the environmental footprint associated with high-performance computing, and this needs to be addressed in future work. 

\section*{Acknowledgements}
This work was supported by the BMBF projects SustainML (Grant 101070408), Albatross (Grant 01IW24002) and by Carl Zeiss Foundation through the Sustainable Embedded AI project (P2021-02-009).

\bibliographystyle{splncs04}
\bibliography{refs}
\end{document}